\documentclass[letterpaper, 10 pt, conference]{ieeeconf}  

\IEEEoverridecommandlockouts                              

\usepackage{graphics} 
\usepackage{amsmath} 
\usepackage{amssymb}  

\usepackage{algorithm} 
\usepackage{algpseudocode}

\usepackage{graphicx}
\usepackage{hyperref}
\usepackage{booktabs}
\usepackage{flushend}
\usepackage{multirow}
\usepackage{makecell}

\usepackage{xcolor}

\title{\LARGE \bf
Improving Safety in Deep Reinforcement Learning \\ using Unsupervised Action Planning
}

\author{
Hao-Lun Hsu$^{1}$, Qiuhua Huang$^{2}$, Sehoon Ha$^{1}$ 
\thanks{$^{1}$ Georgia Institute of Technology, Atlanta, GA, 30308, USA}
\thanks{$^{2}$ Pacific Northwest National Laboratory, Richland, WA, 99352, USA}
\thanks{Emails: {\tt\small hhsu61@gatech.edu, qiuhua.huang@pnnl.gov, sehoonha@gatech.edu}}
}

\begin{document}

\maketitle
\thispagestyle{empty}
\pagestyle{empty}


\newcommand{\haolun}[1]{\textcolor{blue}{{haolun: #1}}}
\newcommand{\sehoon}[1]{\textcolor{magenta}{{Sehoon: #1}}}

\newcommand{\revised}[1]{\textcolor{blue}{#1}}

\newcommand{\original}[1]{\textcolor{magenta}{Original: #1}}
\newcommand{\eqnref}[1]{Equation~(\ref{eq:#1})}
\newcommand{\figref}[1]{Figure~\ref{fig:#1}}
\newcommand{\tabref}[1]{Table~\ref{tab:#1}}
\newcommand{\secref}[1]{Section~\ref{sec:#1}}

\long\def\ignorethis#1{}
\newcommand{\myparagraph}[1]{\noindent\textbf{{#1}}}

\newcommand{\etal}{{\em{et~al.}\ }}
\newcommand{\eg}{e.g.\ }
\newcommand{\ie}{i.e.\ }

\newcommand{\figtodo}[1]{\framebox[0.8\columnwidth]{\rule{0pt}{1in}#1}}



\newcommand{\pdd}[3]{\ensuremath{\frac{\partial^2{#1}}{\partial{#2}\,\partial{#3}}}}

\newcommand{\mat}[1]{\ensuremath{\mathbf{#1}}}
\newcommand{\set}[1]{\ensuremath{\mathcal{#1}}}

\newcommand{\vc}[1]{\ensuremath{\mathbf{#1}}}
\newcommand{\vEndEff}{\ensuremath{\vc{d}}}
\newcommand{\vRelMove}{\ensuremath{\vc{r}}}
\newcommand{\sSet}{\ensuremath{S}}

\newcommand{\vControl}{\ensuremath{\vc{u}}}
\newcommand{\vPoint}{\ensuremath{\vc{p}}}
\newcommand{\sSpringCoef}{{\ensuremath{k_{s}}}}
\newcommand{\sDamperCoef}{{\ensuremath{k_{d}}}}
\newcommand{\vHandle}{\ensuremath{\vc{h}}}
\newcommand{\vForce}{\ensuremath{\vc{f}}}

\newcommand{\mTransChain}{\ensuremath{\vc{W}}}
\newcommand{\mRotateTrans}{\ensuremath{\vc{R}}}
\newcommand{\sJoint}{\ensuremath{q}}
\newcommand{\vJoint}{\ensuremath{\vc{q}}}
\newcommand{\mJoint}{\ensuremath{\vc{Q}}}
\newcommand{\mMass}{\ensuremath{\vc{M}}}
\newcommand{\sMass}{\ensuremath{{m}}}
\newcommand{\vGravity}{\ensuremath{\vc{g}}}
\newcommand{\vConstr}{\ensuremath{\vc{C}}}
\newcommand{\sConstr}{\ensuremath{C}}
\newcommand{\vCOM}{\ensuremath{\vc{x}}}
\newcommand{\sGeneralForce}[1]{\ensuremath{Q_{#1}}}
\newcommand{\vStateVar}{\ensuremath{\vc{y}}}
\newcommand{\vControlVar}{\ensuremath{\vc{u}}}
\newcommand{\tr}[1]{\ensuremath{\mathrm{tr}\left(#1\right)}}

%
%

\renewcommand{\choose}[2]{\ensuremath{\left(\begin{array}{c} #1 \\ #2 \end{array} \right )}}

\newcommand{\gauss}[3]{\ensuremath{\mathcal{N}(#1 | #2 ; #3)}}

\newcommand{\pctab}{\hspace{0.2in}}
\newenvironment{pseudocode} {\begin{center} \begin{minipage}{\textwidth}
                             \normalsize \vspace{-2\baselineskip} \begin{tabbing}
                             \pctab \= \pctab \= \pctab \= \pctab \=
                             \pctab \= \pctab \= \pctab \= \pctab \= \\}
                            {\end{tabbing} \vspace{-2\baselineskip}
                             \end{minipage} \end{center}}
\newenvironment{items}      {\begin{list}{$\bullet$}
                              {\setlength{\partopsep}{\parskip}
                                \setlength{\parsep}{\parskip}
                                \setlength{\topsep}{0pt}
                                \setlength{\itemsep}{0pt}
                                \settowidth{\labelwidth}{$\bullet$}
                                \setlength{\labelsep}{1ex}
                                \setlength{\leftmargin}{\labelwidth}
                                \addtolength{\leftmargin}{\labelsep}
                                }
                              }
                            {\end{list}}
\newcommand{\newfun}[3]{\noindent\vspace{0pt}\fbox{\begin{minipage}{3.3truein}\vspace{#1}~ {#3}~\vspace{12pt}\end{minipage}}\vspace{#2}}

\newcommand{\key}{\textbf}
\newcommand{\fun}{\textsc}



\begin{abstract}

One of the key challenges to deep reinforcement learning (deep RL) is to ensure safety at both training and testing phases. In this work, we propose a novel technique of unsupervised action planning to improve the safety of on-policy reinforcement learning algorithms, such as trust region policy optimization (TRPO) or proximal policy optimization (PPO). We design our safety-aware reinforcement learning by storing all the history of ``recovery'' actions that rescue the agent from dangerous situations into a separate ``safety'' buffer and finding the best recovery action when the agent encounters similar states. Because this functionality requires the algorithm to query similar states, we implement the proposed safety mechanism using an unsupervised learning algorithm, k-means clustering. We evaluate the proposed algorithm on six robotic control tasks that cover navigation and manipulation. Our results show that the proposed safety RL algorithm can achieve higher rewards compared with multiple baselines in both discrete and continuous control problems. The supplemental video can be found at: https://youtu.be/AFTeWSohILo.

\end{abstract}

\section{Introduction} \label{sec:intro}
Deep reinforcement learning (deep RL) has emerged as a promising approach for developing highly intelligent agents from simple formulations.
One of the biggest challenges for deploying deep RL trained agents to real-world autonomous applications is its safety issue because a failure of a robotic agent can cause costly damage to the robot and its surroundings, including nearby humans.
However, it is very challenging to guarantee the safety of the RL agents due to their poor extrapolation capabilities.
In addition, the safety of a robotic agent is not only important at the testing phase.
Many researchers have demonstrated that the learning-on-real-robot philosophy \cite{kalashnikov2018qt,ha2020learning} can lead to a state-of-the-art real-world performance by directly collecting experience from the real world and intrinsically bypassing the reality gap.
Therefore, it is important to develop a safe reinforcement learning algorithm that can minimize the number of failures at both training and testing time.

To this end, the safety issue has been considered an important problem in the deep RL community \cite{conservative, SQRL, intrinsic, risk-sens, cross-entropy}. Once the agent can predict the risk of the state, it can plan accordingly to avoid failures by using planning algorithms, such as model-predictive control  (MPC) \cite{slippery,Multiconstraints, LBMPC} or rejection sampling \cite{weak, implicit, roadmap}.
However, our goal in this paper is to model a little bit different mechanism to avoid risky situations. When humans encounter a dangerous situation, we often try to recall the past experience of overcoming similar situations and try to avoid the risk by repeating similar recovery actions.
This kind of mechanism may make the exploration conservative and slow down the learning a bit, but it is expected to be helpful to reduce the number of failures.

\begin{figure}
\centering
\includegraphics[scale=0.6]{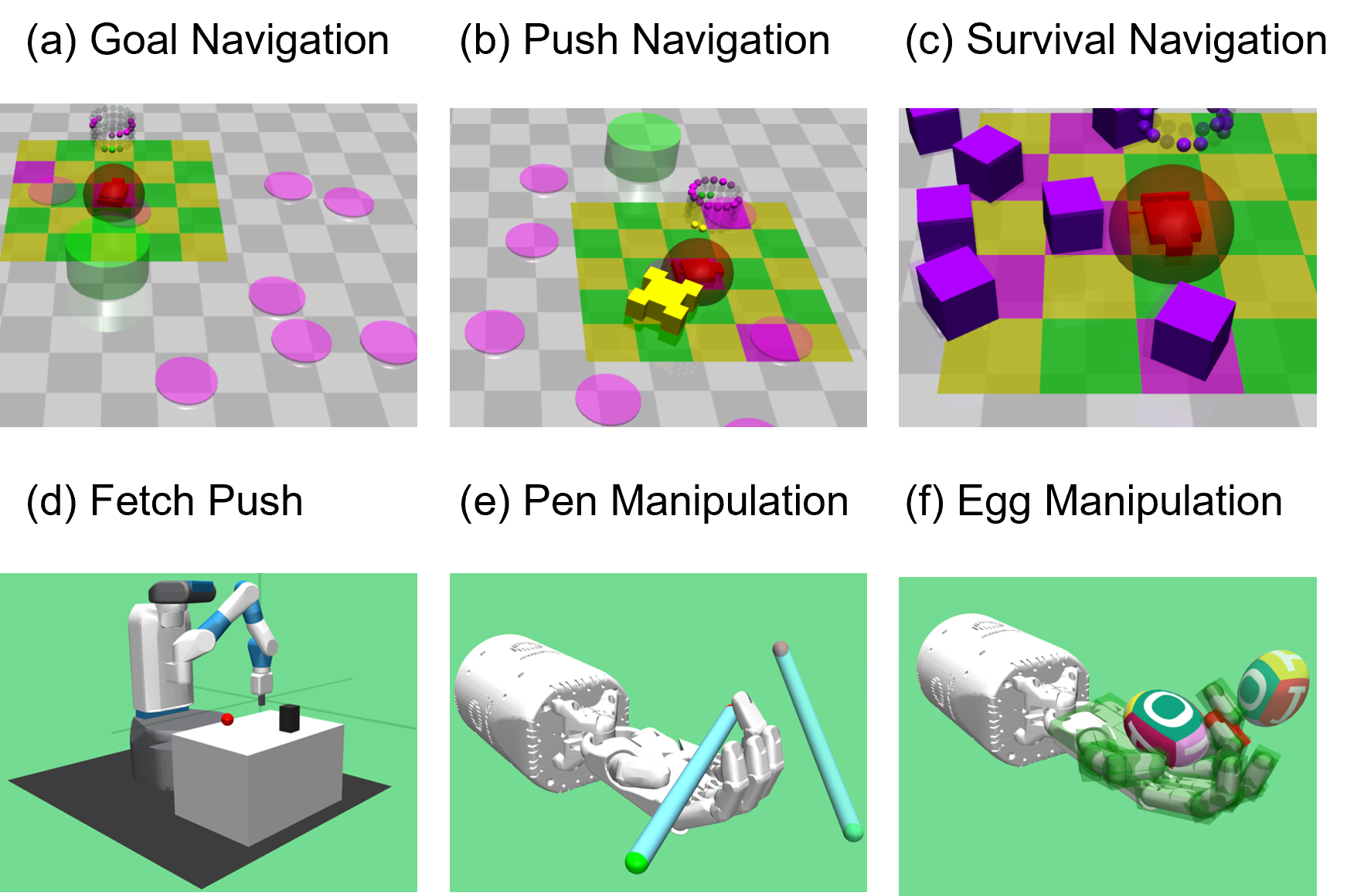}
\caption{Illustrations of six environments evaluated in our approach, including navigation and manipulation tasks. Our method is generalizable enough to be deployed in a variety of environments. 
}
\label{env_task}

\end{figure}

In this paper, we propose a novel technique of unsupervised action planning to improve the safety of on-policy reinforcement learning algorithms.
When the agent is in dangerous states, our algorithm searches for the best ``recovery’’ action in history, which allows the agent to recover from similar risky states and get back to a safe area.
Because this mechanism requires the algorithm to search for similar states and find the best action, we implement such a safe action planning mechanism using an unsupervised learning algorithm, k-means clustering \cite{classification, Heuristic, kmeans}.
Our safety buffer stores all the successful recovery actions and organizes all the recovery actions by using a k-means clustering algorithm so that an RL algorithm can efficiently search for similar recovery actions in history. 

We benchmark our algorithm in six environments using the MuJoCo physics engine \cite{mujoco} within the OpenAI Safety Gym \cite{benchmark_safe} and Gym \cite{benchmark_gym} frameworks illustrated in Figure~\ref{env_task}. These environments cover a variety of robotic control tasks, such as navigating with collision avoidance, pushing with a Fetch robotic arm, and in-hand object manipulation.
Our safety action planning algorithm allows us to successfully reduce the number of failures and eventually obtain significantly higher rewards at the end.

Our technical contributions are as follows:

\begin{itemize}
    \item We propose a novel safety-aware reinforcement learning using unsupervised action learning.
    \item We demonstrate that the proposed technique can successfully reduce the number of failures at both training and testing phases for six problems in robotics.
\end{itemize}

\section{Related Work}
\label{sec:rel}

It is crucial to guarantee the safety of the agent when we deploy RL-trained policies on real-world applications. Many existing techniques for incorporating the safety issue can be roughly categorized by two approaches \cite{comprehensive}. The first is based on the modification of the optimization criterion with a safety factor. The second is based on the modification of the exploration process by incorporating the external knowledge or the guidance of an explicit risk metric. 
In the first category, one notable approach is to formulate safety in RL as a constrained Markov Decision Process (CMDP) \cite{cmdp}. However, Ray et al.~\cite{benchmark_safe} demonstrated that constrained RL algorithms do not always satisfy the given constraints, especially at the beginning of the training process. Conservative exploration via prior knowledge \cite{comprehensive}  can be an alternative approach to reduce undesirable situations, such as the explicit awareness of danger, a data-driven based approach to model the risk, or both. 

\subsection{Constrained policy optimization}
One common method to solve CMDPs is to utilize the Lagrangian method \cite{risk} with any policy gradient (PG) algorithms. The original RL objective is penalized with constraint violations before computing the saddle point of the constrained policy optimization by primal-dual methods \cite{cmdpLag}. Due to the conflicting goals of maximizing cumulative rewards and minimizing failure, Lagrangian approaches only ensure safety asymptotically and may lead to danger during training \cite{lyapunov}. Constrained policy optimization (CPO) \cite{cpo}, extends trust-region policy optimization (TRPO) \cite{trpo} to solve CMDPs for satisfying constraints during both training and testing phases. Although CPO has shown promising empirical results on high-dimensional constrained control tasks, approximation errors in CPO prevent it from fully satisfying constraints in more complicated environments \cite{benchmark_safe}. Chow et al.~\cite{lyapunov} presented $\theta$-projection based on a Lyapunov function approach which provides a more general framework for both on-policy and off-policy learning algorithms. Sikchi et al.~\cite{barrier} analyzed that $\theta$-projection still relies mostly on backtracking to ensure safety rather than safe projection, which explains why it might not limit unsafe behaviors effectively in some scenarios. Yang et al.~\cite{projection} developed Projection-based Constrained Policy Optimization (PCPO), which projects an intermediate policy onto the constraint set, to provide a certain amount of safety guarantees. Yang et al.~\cite{accelerating} further extended PCPO by updating the policy from a baseline policy with dynamic distances. 

\subsection{Conservative exploration}
In order to learn and to avoid the regions where constraint violations are likely to happen, many researchers have variously investigated for modeling catastrophic or near-catastrophic failures. For example, some previous algorithms \cite{guarantee,predictive} have the access to a default safe policy and a safe set of environment states. Datal et al.~\cite{dalal2018} integrated prior knowledge to project the action from the policy onto a safe set. Bharadhwaj et al.~\cite{conservative} proposed conservative safety critics for estimating the probability of failure and resampling actions. Introducing external models can also modify the progress of its exploration: for instance, the work of updating actor and advisor policies together to minimize total cost and risk constraint violation \cite{dynamic}.  Turchetta et al.~\cite{curriculum} provided the agent with alternative actions via a teacher policy to save the agent from violating constraints. The external models are not limited in RL policies. For example, danger index can be predicted with Negative-Avoidance Function \cite{autonomous} and intrinsic fear is approximated by a supervised learning model \cite{intrinsic}, which all can intervene an original action and improve the safety. Researchers \cite{recovery,trace,multi} also have investigated reset policies, which are trained jointly with the task policy to gradually expand recoverable regions. 
In this work, we introduce an unsupervised learning mechanism to directly remember recovery actions, which is compatible with any on-policy (PG) algorithms and can improve high failure rate of the existing Lagrangian methods at the early training. 

\section{Safe Reinforcement Learning via Unsupervised Action Planning} \label{sec:method}
\begin{figure}
\centering
\includegraphics[scale=0.36]{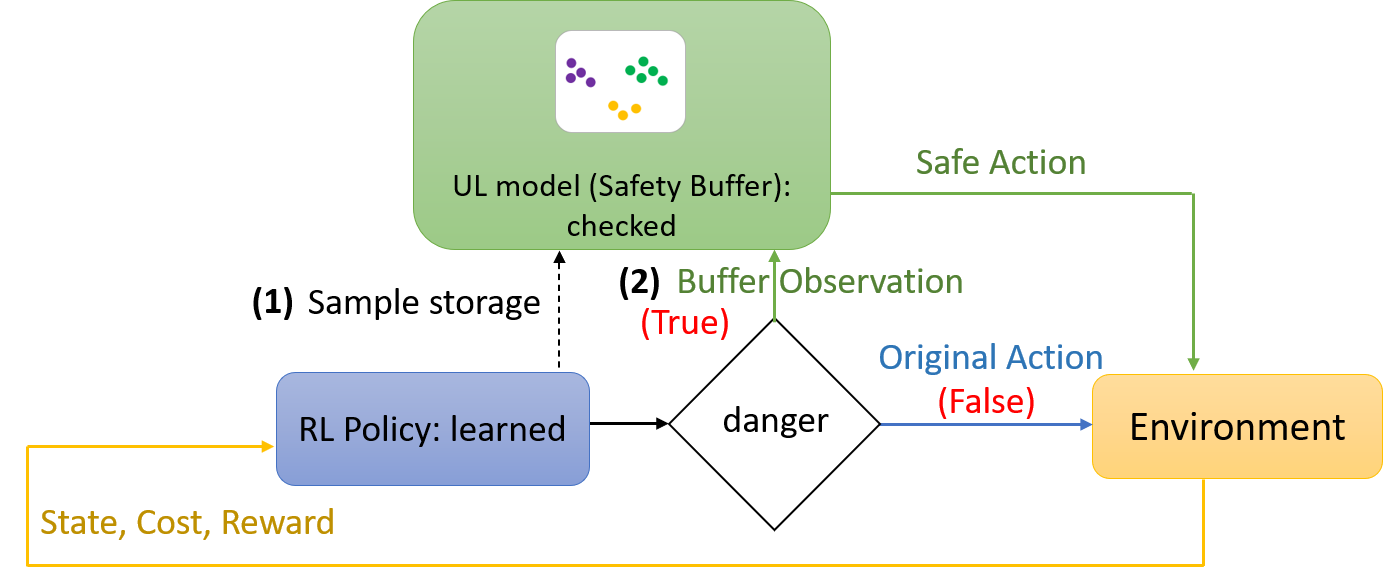}
\caption{An overview diagram of the proposed safe RL framework with unsupervised action planning.
We store all the ``recovery'' actions that allow the agent to escape from dangerous states into a separate safety buffer and find the best recovery action when it encounters a similar situation.
For an efficient query, we manage the safety buffer using unsupervised learning, k-means clustering.
}
\label{framework_diagram}

\end{figure}

In this section, we will present our safe reinforcement learning algorithm that achieves conservative exploration via unsupervised action planning. 
Intuitively, our algorithm stores all the previous ``recovery'' actions, which successfully reduce the risk of the agent, into a separate replay buffer and tries to find the best recovery action when the agent encounters similar dangerous situations.
Because this ``search-and-max'' operation is not intuitive to learn using standard supervised learning algorithms, we implement this mechanism using unsupervised learning, k-means clustering.
The overview of the proposed method is illustrated in Figure~\ref{framework_diagram}.

\begin{algorithm*}
	\caption{Safe RL using Unsupervised Action Planning} \label{alg:main}
	\begin{algorithmic}[1]
	    \State Initialize policy $\pi_{\phi}$ and safety buffer $D$
	    \State Pre-train the policy $\pi$ for a small number of epochs
	  
		\For {$epoch=1,2,\ldots$}
		    
		    \State $[s_{0}, c_{0} ]\sim P(s_{0}, c_{0})$ \Comment{Initialize state $s$ and cost $c$}\label{line:3}
			\For {$t=0,1,\ldots,T$}
			   
			    \State $a_{t} \sim \pi_{\phi}(a_{t}| s_{t}) $ \label{line:6}
			    \State $b_t = b(s_t)$ \Comment{Extract the state features}
			    
			    \If {$c_{t} \geq \hat{c}$}\label{line:8} \Comment{If dangerous}
			        \State $a_{t}$ = queryRecoveryAction($a_{t}, b_{t}$, $D$)\label{line:9} \Comment{Activate safety protection mechanism}
			    \EndIf \label{line:10}
			    \State $[s_{t+1}, c_{t+1}, b_{t+1}, r_{t} ]\sim P(s_{t+1}, c_{t+1}, b_{t+1}, r_{t} | s_{t}, a_{t})$ 
			   \If {$c_{t} \geq \hat{c}$ and $c_{t+1} < \hat{c}$} \Comment{If recovers from danger} \label{line:12}
			        
		            \State $D \leftarrow\ D \cup {(b_{t}, a_{t}, r_{t})}$ \label{line:11}
			       
			    \EndIf
			    
			    \State $s_{t} \leftarrow\ s_{t+1},\ c_{t}\leftarrow\ c_{t+1},\
			    b_{t}\leftarrow\ b_{t+1}$

			    \If {end of the episode} 
			        \State Rebuild clusters in the safety buffer $D$ \label{line:17}
			       \Comment{Regularly updates clusters}
	           \EndIf

			\EndFor

		    \State Update $\pi_{\phi}$ \Comment{Standard RL steps}
		\EndFor
	\end{algorithmic}
\end{algorithm*}

\begin{algorithm}
	\caption{queryRecoveryAction}\label{alg:safeAction}
	\begin{algorithmic}[1]
	    \State \textbf{Input:} action $a_{t}$, state feature $b_{t}$, and the safety buffer $D$
	    \State Acquire an action set $A$ containing actions in the same cluster with $b_{t}$
	    \If{$a_{t} \in A$}
	        \State \textbf{return} $a_{t}$
	   \Else
	        \State \textbf{return} the action $\tilde{a_{t}}\in A$ with the maximum reward
	   \EndIf
	   
	\end{algorithmic} 
\end{algorithm}

\subsection{Problem Formulation} \label{sec:problem}

A standard RL task can be formulated as a Markov Decision Process (MDP) defined by a tuple ($S, A, r, T, P$), where $S$ and $A$ are state and action spaces, $r$ is a reward function, $T$ is the set of terminal conditions, and $P$ is the transition probability. 
The goal of reinforcement learning is to find the optimal policy $\pi$ maximizing the accumulated reward, typically with the discount factor $\gamma$.
The Constrained MDP (CMDP) model extends MDP by introducing additional costs, defined by ($S, A, r, T, P, C$), where $C$ is a constraint cost function. 
We consider a single constraint $c(s_{t})$ (or simply $c_t$) in each step, where $c(s_{t}) = 0$ when a state is safe, $c(s_{t}) = 1$ when a state fails, and some in-between value when a state is in danger. 
Although many safety RL algorithms adopt binary success/fail flags, we can measure risk as a continuous value in many different robotic control tasks. For instance, we can quantify the risk as to the distance to the nearest object using LiDAR in the autonomous driving task. For manipulation, we can estimate the position and orientation of the object using the off-the-shelf pose estimator~\cite{PoseRBPF} and compute its tilting angle to determine whether it is about to fall or not.

With the continuous cost function, our goal is to minimize the number of failures ($c(s_{t}) = 1$) instead of danger states. The intermediate cost values are warning signals for the agent instead of the cost to be minimized. Therefore, without loss of generality, we consider the state is in danger if $c(s_{t})$ is greater than the threshold, $\hat{c}$, which is set to 0.5 for all experiments.

In addition, we assume a feature function $b(s)$ that extracts features from the state $s$, or in short notation, $b_t = b(s_t)$. This feature is used to store previous experience into a separate ``safety'' buffer, which is managed by unsupervised learning.
Typically, we take the subset of the relevant observation channels to obtain a compact feature representation. For more details, please refer to the experimental section.

\subsection{Safe Action Planning with Unsupervised Learning} 
We propose to augment a standard RL framework with conservative exploration by planning safe actions via unsupervised learning.
We modify the existing RL algorithm by storing ``recovery'' actions that bring the agent from a danger area to a safe area into a separate ``safety'' buffer and searching for the best recovery action of the similar dangerous states.
Therefore, our algorithm must be able to 1) look up similar previous history and 2) find the most successful recovery action.

\noindent \textbf{Safety buffer.} 
To this end, we introduce an additional data structure, the safety buffer $D$, to organize all the historical successful recovery actions that are defined by: $c_{t} \geq \hat{c}$ and $c_{t+1} < \hat{c}$, where $c_t = c(s_t)$.
Please note that this is different from a regular replay buffer that is flushed at each learning step of on-policy learning.
In our definition, the safety buffer $D$ stores a tuple of ($b_t, a_t, r_t$) and clusters all the tuples in the feature space $b_t$.
Therefore, the safety buffer $D$ allows us to find the best action in similar states by comparing the associated rewards in the same cluster, i.e., all the tuples with similar features.
We maintain the clusters up-to-date by rebuilding clusters per every episode.

\noindent \textbf{Action filtering.}
When the policy generates an action $a_t$, we first check whether the current state $s_t$ is dangerous or not.
If the state is considered to be risky ($c_{t} \geq \hat{c}$), we look up the safety buffer $D$ to find the best recovery action as follows.
First, our algorithm acquires a candidate action set $A$, which contains all the actions in the same cluster with the feature of the current state $b_t$.
If the current action $a_t$ is in a candidate action set $A$, the algorithm considers it a recovery action and executes it as is. 
If not, the algorithm chooses an alternative action $\tilde{a}_t$ from the candidate set $A$ that is associated with the maximum reward $r_t$.

\noindent \textbf{Learning procedure.}
We select an on-policy RL algorithm as a backbone of learning although the basic concept of our approach can be general to any RL algorithms.
Figure~\ref{framework_diagram} and Algorithm~\ref{alg:main} illustrate the augmented learning process while Algorithm~\ref{alg:safeAction} shows the action planning process.
Line \ref{line:6}-\ref{line:10} in Algorithm~\ref{alg:main} is our conservative exploration mechanism via safe action planning, which is previously described.
If the action is verified as a recovery action (line \ref{line:12}), we add the experience to the safety buffer $D$.
At the end of every episode, we rebuild the clusters to keep the safety buffer up-to-date (line \ref{line:17}).
Once all the trajectories are collected, we invoke the standard update step of the on-policy RL algorithm.

\begin{figure*}
    \centering
    \includegraphics[width=1\linewidth]{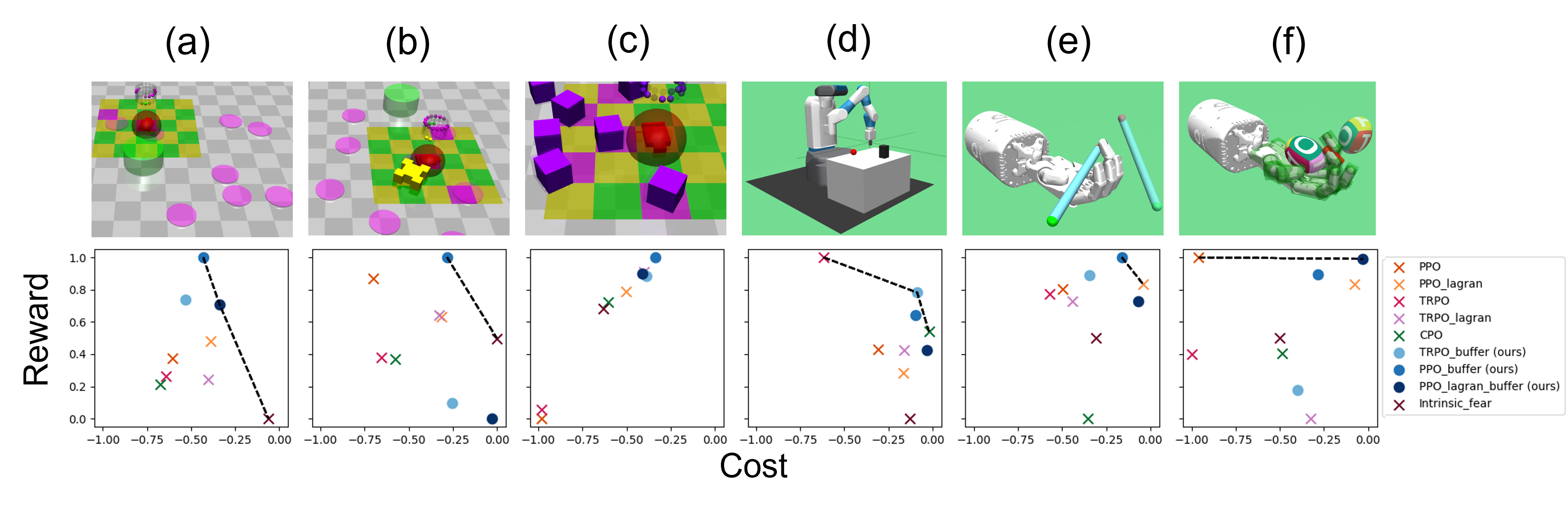}
    \caption{Illustrations of the six environments in our experiments (\textbf{Top}) and their corresponding performance (\textbf{Bottom}). We highlight the Pareto optimal solutions by connecting with dotted lines. (a) Goal achieving in navigation while avoiding static hazards. (b) Pushing in navigation while avoiding static hazards. (c) Survival from moving obstacles. (d) Fetch push without toppling over. (e) In-hand pen manipulation without falling off.  (f) In-hand egg manipulation without crush. We evaluate the performance of our approaches and the baselines via the reward and average cost.
    The proposed algorithm (blue-ish dots) shows good performance in terms of both average rewards and costs.
    }
    \label{env_performance}
\end{figure*}

\begin{figure}
    \centering
    \includegraphics[width=1.02\linewidth]{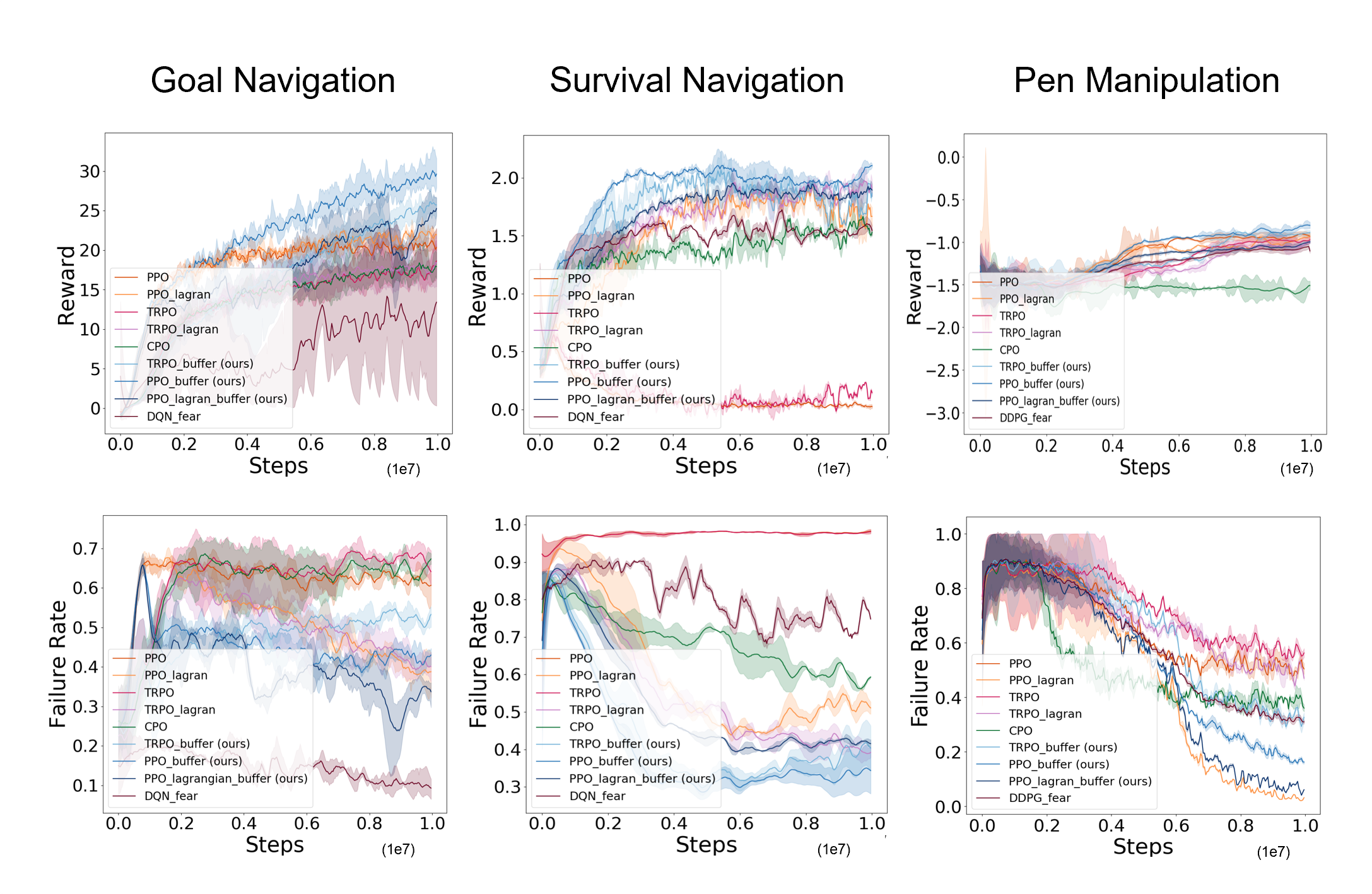}
    \caption{Learning curves of our approaches (blue) and baselines averaged over 3 seeds for three robotic control tasks. \textbf{Top:} average task rewards (higher is better). \textbf{Bottom:} average failure rate (lower is better). Our approaches remedy the high failure rate at the early training of the Lagrangian approaches while achieving higher rewards. 
    }
    \label{learning_curve}
\end{figure}

\section{Experiments}
We design experiments to validate that the proposed system can successfully reduce the number of failures for both training and test phases while achieving higher rewards. Particularly, our goal is to investigate the following questions:

\begin{enumerate}
  \item Can the proposed technique improve the training and testing time safety of on-policy learning algorithms, such as PPO \cite{ppo} and TRPO \cite{trpo}, compared to other baseline algorithms?
  \item How does the hyperparameters, such as the number of the clusters or the danger threshold, affect the performance of the algorithms?



\end{enumerate}

\subsection{Benchmark Problems}

Our algorithm is designed to be general, and we aim to empirically evaluate the proposed technique on six different robotic control tasks. We start with discrete action space in 2D navigation tasks so that the agent can directly check whether the current action is in a candidate action set from the safety buffer. However, we further extend our approach using a discretized grid-bucket to identify approximate action for continuous action space in the tasks via a Fetch robotic arm and in-hand object manipulation. We describe each task in this section and show the illustrations in Figure \ref{env_performance}.

\myparagraph{Tasks with discrete action space.} We firstly modify Safety Gym environments to three different scenarios for navigation. The agent must either achieve or push a box to a goal position while avoiding static traps in the first two tasks. In the Survival Navigation task, the agent should learn to survive from dynamic and moving traps. The agent moves on a 2D map with discrete actions while it fails due to the collision with any trap (e.g., hazard or obstacle). We utilize rasterized feature matrix from the map surrounding the agent for collision avoidance.

\myparagraph{Tasks with continuous action space.}  In the Fetch Push task, the agent aims to push a vertically rectangular block to a goal position using a 7-DoF Fetch arm without the block toppling over. One task from in-hand object manipulation is to manipulate a pen to achieve a target pose without falling off. Because the failures are defined as the case when the block toppling over in the Fetch Push task and when the pen falls off in the Pen Manipulation task, we extract all z-axis related information along z-axis (e.g., position and velocity) of the object to define safety features. The other scenario we test our method in the hand manipulation task is to replace a pen with an egg. Since an egg is easily crushed, we incorporate the values of contact normal forces from sensors to the original states. When any contact normal force is larger than a specific value ($20$~N in our experiment), the task is seen as a failure. We store the values from sensors as features into the safety buffer.


\begin{table*} 
\begin{center}
\caption{\label{tab:cumulative_failure}Relative Cumulative Failures During Learning}   
\begin{tabular}{cccccccccc}    

 \cmidrule {1 -10} \textbf{Task/ Algorithm} & PPO & TRPO & PPO+Lag & TRPO+Lag & CPO & \makecell{TRPO+Buffer\\(Ours)} &  \makecell{PPO+Lag\\(Ours)} &  \makecell{PPO+Lag+Buff\\(Ours)} & Intrinsic Fear\\    \midrule  
 Goal Navigation & 1.00 & 0.90 & 0.70 & 0.39 & 0.58 & 0.37 & 0.43 & 0.29 & \textbf{0.14}\\  
 Push Navigation & 0.99 & 1.00 & 0.51 & 0.46 & 0.74 & 0.22 & 0.32 & 0.15 & \textbf{0.10}\\ 
 Survival Navigation & 1.00 & 0.98 & 0.15 & 0.13 & 0.17 & 0.06 & \textbf{0.05} & 0.12 & 0.70\\ 
 Fetch Push w/o Toppling & 0.72 & 1.00 & 0.29 & 0.32 & 0.06 & 0.16 & 0.13 & \textbf{0.04} & 0.19\\ 
 Pen Manipulation w/o Falling & 0.98 & 1.00 & 0.41 & 0.76 & \textbf{0.36} & 0.56 & 0.45 & 0.41 & 0.50\\ 
 Egg Manipulation w/o Crush & 1.00 & 0.90 & 0.15 & 0.39 & 0.27 & 0.21 & 0.13 & \textbf{0.12} & 0.33\\
 \bottomrule    
\end{tabular}  

\end{center}
\end{table*}

\begin{table*} 
\begin{center}
\caption{\label{tab:clusters}Performance of our approach with different number of clusters. In our notation, $N^{1/2}$ refers
to the square root and $N^{1/3}$ is the cube root of the total number of samples. Each entry in the table
indicates the corresponding reward (R) and failure rate (F).} 
\begin{tabular}{lllllllllll}    
\toprule & \multicolumn{2}{c}{Brute Force} & \multicolumn{2}{c}{$N^{1/10}$} & \multicolumn{2}{c}{$N^{1/3}$} & \multicolumn{2}{c}{$N^{1/2}$} & \multicolumn{2}{c}{$N^{8/10}$}\\
 \cmidrule {2-11} \textbf{Task/ Number of Clusters}  & R & F & R & F  & R & F & R & F  & R & F\\    \midrule  
 Goal Navigation & \textbf{28.52} & 0.47 & 20.66 & 0.56  & 25.37 & \textbf{0.43} & 28.44 & 0.45 & 25.32  & 0.44\\  
 Push Navigation & 2.69 & 0.30 & 2.41 & 0.38  & \textbf{3.12}  & 0.35 & 2.93 & 0.31 & 2.72 & \textbf{0.29} \\
 Survival Navigation & 1.73 & 0.33 & 1.06 & 0.37  & 1.95  & 0.32 & \textbf{2.03} & \textbf{0.31} & 1.55 & 0.36 \\
 Fetch Push w/o Toppling & -0.39 & 0.08 & -0.41 & \textbf{0.05}  & -0.40  & 0.08 & \textbf{-0.32} & 0.10 & -0.37 & 0.11 \\  
 Pen Manipulation w/o Falling & -1.00 & 0.24 & -1.20 & 0.37  & -1.13  & 0.24 & -0.87 & \textbf{0.21} & \textbf{-0.84} & 0.26 \\ 
 Egg Manipulation w/o Crush & -1.84 & 0.33 & -1.91 & 0.35 & \textbf{-1.77} & 0.29  & -1.82  & \textbf{0.25} & -1.80 & 0.28 \\
 \bottomrule   
  
\end{tabular}  

\end{center}
\end{table*}

\begin{figure}
    \centering
    \includegraphics[width=1.0\linewidth]{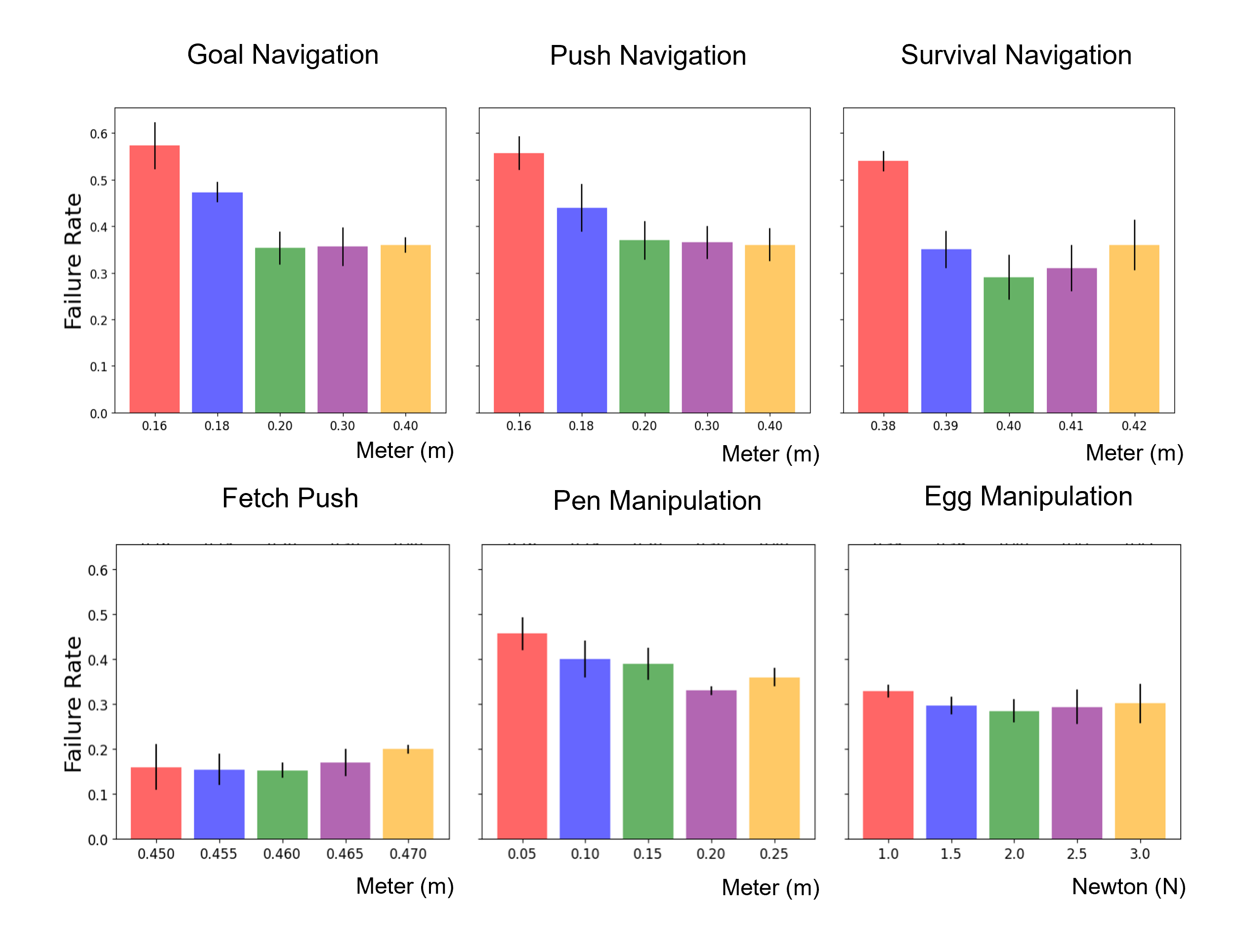}
  
    \caption{Testing results with ablation studies in different danger thresholds (states with $c(s_t) \geq  \hat{c}$) for all robotic control tasks. For all subplots, smaller x values indicate more aggressive policies while being more conservative with larger x values.
    }
    \label{threshold_ablation}
\end{figure}

  


\subsection{Comparison against Baselines}
To demonstrate the advantage of augmenting reinforcement learning with unsupervised learning for safe actions, we compare our algorithms with a set of baseline algorithms. We select the default baseline algorithms that are implemented in the original Safety Gym environment \cite{benchmark_safe}, including PPO and TRPO in both their original unconstrained forms (\emph{PPO}, \emph{TRPO}) and the augmented formulations (\emph{PPO-lagran}, \emph{TRPO-lagran}) with adaptive penalties for safety costs based on the Lagrangian approach to solve constrained optimization, as well as CPO, as our baseline.  We also validate how the difference in the definition of danger states influences the exploration via intrinsic-fear DQNs \cite{intrinsic}, which incorporates DQN with a supervised risk-assessment model to guide exploration process to avoid undesirable behaviors. Instead of defining a fixed radius of the danger area, the states within $k_r$ steps before failure are seen danger. To deal with continuous action space, we adapt intrinsic-fear to DDPG. We demonstrate three versions of our algorithm that augment the baseline algorithms mentioned above. The first and the second are \emph{PPO-buffer} and \emph{TRPO-buffer} that augment vanilla algorithms with the proposed safety mechanism. The last one is \emph{PPO-lagran-buffer} that combines \emph{PPO-lagran} with the safety buffer approach.

Figure \ref{learning_curve} shows the learning curves of three tasks as examples. To have a better interpretation on compact results, we normalize the reward and scale the value of failure rate to between $-1$ and $0$, called \emph{average cost}, in Figure \ref{env_performance}. The closer the average cost to $0$, the less the failure rate it is. In general, our algorithm (\emph{PPO-buffer}, \emph{TRPO-buffer}, \emph{PPO-lagran-buffer}, all represented in blue-ish colors) can achieve higher or matching rewards while significantly reducing the number of failures, even compared to the baseline algorithms (\emph{PPO-lagran}, \emph{TRPO-lagran}, \emph{CPO}) that are designed to solve constrained MDP problems.
Our approaches remedy the high failure rate at the early training of the baseline methods, and \emph{PPO-buffer} has the best trade-off between the reward and failure rate. We observe the generalizability of our approach \emph{PPO-buffer} in all the tasks, which other baselines cannot achieve. For example, \emph{CPO} gets the lowest reward in the Pen Manipulation (without falling) task, and \emph{PPO-lagran} has the lower rewards compared with \emph{PPO-buffer} among all tasks. In addition, since the states within $k_r$ steps before failure are all seen as danger states in \emph{Intrinsic-fear}, the agent misunderstands that the danger states will always result in failure, which makes the agent learn conservatively. Therefore, \emph{Intrinsic-fear} can get the minimum failure rate in the goal navigation task while the reward is also the lowest.


As we discussed earlier, the safety is not only important at testing but also crucial for training in some scenarios, such as learning or fine-tuning on real-worlds.
To this end, we also compare the relative cumulative number of failures during the training in Table~\ref{tab:cumulative_failure}.
The results show that we can successfully reduce the cumulative number of failures by adding the proposed safety mechanism in most of the tasks while \emph{Intrinsic-fear} accumulates lots of failures to learn the Survival Navigation task. 
Our approach leads to the least number of failures in the Egg Manipulation (without crush) task, indicateing that our approach can be scaled to solve the problem in a high-dimensional continuous action space.


\subsection{Analysis} \label{sec:4.4} 
We conduct additional experiments and analyze the failure rate with respect to different choices of the number of clusters for k-means clustering and the danger threshold in \emph{PPO-buffer} in this section. 
 

\myparagraph{Number of clusters. }
First, we investigate the performance of the proposed algorithm with respect to the different number of clusters (Table~\ref{tab:clusters}). Initially, we hypothesize that acquiring an ``enough’’ number of clusters leads to a good performance. In general, we observe that the choice of $N^{1/2}$ shows the reasonable performance for most of the tasks, where $N$ is the number of samples stored into the safety buffer, while their peak values can differ for different scenarios. To this end, we conduct all the other experiments with the $N^{1/2}$ clusters.

\myparagraph{Danger Threshold. }
In our definition, a danger threshold is a predefined value where an agent starts to receive a positive constraint value, $c(s_t) \geq \hat{c}$, where $\hat{c} = 0.5$. In all navigation tasks, the danger thresholds are the radius of a circling zone starting from any trap. The agent is dangerous whenever it enters the danger zone. The danger thresholds in both Fetch Push and Pen Manipulation (without falling) tasks are defined along the z-axis so that the danger can be told from the height of the agent's center of mass. We define the margin value between $20$~N (failure) and the largest contact normal force as the danger threshold in the Egg Manipulation (without crush) task. For example, $1.0$~N in Figure \ref{threshold_ablation} indicates that the state is in danger when the largest contact normal force is more than $19$~N. If the danger threshold is too small, the agent may not have enough time to be aware of the risk.  If it is too large, the agent may learn too conservative behaviors.

Figure \ref{threshold_ablation} illustrates the trend of failure rates against the selection of the danger threshold. For all subplots, smaller x values indicate more aggressive policies and vice versa. As we expected, most of the tasks show lower failure rates in more conservative settings. One notable exception is the Fetch Push task, where a large threshold cannot capture the tipping moments of the block precisely. However, our approach still earns relative low failure rate compared with all baselines.

\section{CONCLUSION and FUTURE WORK}
\label{sec:conclusion} 
We propose a technique using unsupervised action planning to learn a safe RL policy that is capable of avoiding dangerous situations. We focus on resolving key challenges in minimizing the number of failures during the training phase as well as improving the final performance during the testing phase. We augment a general on-policy RL algorithm with our safety buffer, which organizes recovery actions with unsupervised learning. We exploit recovery actions stored in the safety buffer to bring an agent from a risky state to a safe zone. Our experiments demonstrate that the proposed method successfully reduce the number of failures during training and testing phases, and also allow the agent to achieve higher rewards even in hand manipulation task with high-dimensional action space.

Currently, our algorithm considers only one-step recovery action, which may not be sufficient for some complex situations. In the future, we want to extend our algorithm to support multi-step planning by having a short history of actions to the safety buffer, which can improve the performance of recovery action planning. Another interesting future research direction is to test our algorithm on safety-critical problems, such as medical applications or power grid control.

\bibliography{main}

\begin{thebibliography}{10}
\providecommand{\url}[1]{#1}
\csname url@samestyle\endcsname
\providecommand{\newblock}{\relax}
\providecommand{\bibinfo}[2]{#2}
\providecommand{\BIBentrySTDinterwordspacing}{\spaceskip=0pt\relax}
\providecommand{\BIBentryALTinterwordstretchfactor}{4}
\providecommand{\BIBentryALTinterwordspacing}{\spaceskip=\fontdimen2\font plus
\BIBentryALTinterwordstretchfactor\fontdimen3\font minus
  \fontdimen4\font\relax}
\providecommand{\BIBforeignlanguage}[2]{{%
\expandafter\ifx\csname l@#1\endcsname\relax
\typeout{** WARNING: IEEEtran.bst: No hyphenation pattern has been}%
\typeout{** loaded for the language `#1'. Using the pattern for}%
\typeout{** the default language instead.}%
\else
\language=\csname l@#1\endcsname
\fi
#2}}
\providecommand{\BIBdecl}{\relax}
\BIBdecl

\bibitem{kalashnikov2018qt}
D.~Kalashnikov, A.~Irpan, P.~Pastor, J.~Ibarz, A.~Herzog, E.~Jang, D.~Quillen,
  E.~Holly, M.~Kalakrishnan, V.~Vanhoucke \emph{et~al.}, ``Qt-opt: Scalable
  deep reinforcement learning for vision-based robotic manipulation,''
  \emph{arXiv preprint arXiv:1806.10293}, 2018.

\bibitem{ha2020learning}
S.~Ha, P.~Xu, Z.~Tan, S.~Levine, and J.~Tan, ``Learning to walk in the real
  world with minimal human effort,'' \emph{arXiv preprint arXiv:2002.08550},
  2020.

\bibitem{conservative}
H.~Bharadhwaj, A.~Kumar, N.~Rhinehart, S.~Levine, F.~Shkurti, and A.~Garg,
  ``Conservative safety critics for exploration,'' \emph{International
  Conference on Learning Representations (ICLR)}, 2021.

\bibitem{SQRL}
K.~Srinivasan, B.~Eysenbach, S.~Ha, J.~Tan, and C.~Finn, ``Learning to be safe:
  Deep rl with a safety critic,'' \emph{arXiv preprint arXiv:2010.14603}, 2021.

\bibitem{intrinsic}
Z.~C. Lipton, K.~Azizzadenesheli, A.~Kumar, L.~Li, J.~Gao, and L.~Deng,
  ``Combating reinforcement learning’s sisyphean curse with intrinsic fear,''
  \emph{arXiv preprint:1611.01211}, 2016.

\bibitem{risk-sens}
P.~Geibel and F.~Wysotzki, ``Risk-sensitive reinforcement learning applied to
  control under constraints,'' \emph{Journal of Artificial Intelligence
  Research}, vol.~24, pp. 81--108, 2005.

\bibitem{cross-entropy}
Z.~Liu, H.~Zhou, B.~Chen, S.~Zhong, M.~Hebert, and D.~Zhao, ``Constrained
  model-based reinforcement learning with robust cross-entropy method,''
  \emph{arXiv preprint arXiv:2010.07968}, 2020.

\bibitem{slippery}
Y.~Gao, T.~Lin, F.~Borrelli, E.~Tseng, and D.~Hrovat, ``Predictive control of
  autonomous ground vehicles with obstacle avoidance on slippery roads,''
  \emph{ASME 2010 Dynamic Systems and Control Conference}, vol.~1, pp.
  265--272, 2010.

\bibitem{Multiconstraints}
J.~Ji, A.~Khajepour, W.~W. Melek, S.~Member, and Y.~Huang, ``Path planning and
  tracking for vehicle collision avoidance based on model predictive control
  with multiconstraints,'' \emph{IEEE Transactions on Vehicular Technology},
  vol.~66, no.~2, pp. 952--964, 2016.

\bibitem{LBMPC}
A.~Aswani, H.~Gonzalez, S.~S. Sastry, and C.~Tomlin, ``Provably safe and robust
  learning-based model predictive control,'' \emph{Automatica}, vol.~49, no.~5,
  p. 1216–1226, 2013.

\bibitem{weak}
A.~Ghadirzadeh, P.~Poklukar, X.~Chen, H.~Yao, H.~Azizpour, M.~Björkman,
  C.~Finn, and D.~Kragic, ``Few-shot learning with weak supervision,''
  \emph{ICLR workshop on Learning to Learning}, 2021.

\bibitem{implicit}
C.~Zhang, J.~Huh, and D.~D. Lee, ``Learning implicit sampling distributions for
  motion planning,'' \emph{2018 IEEE/RSJ International Conference on
  Intelligent Robots and Systems (IROS)}, p. 3654–3661, 2018.

\bibitem{roadmap}
V.~Boor, M.~H. Overmars, and A.~F. van~der Stappen, ``The gaussian sampling
  strategy for probabilistic roadmap planners,'' \emph{Proceedings 1999 IEEE
  International Conference on Robotics and Automation}, vol.~2, pp. 1018--1023,
  1999.

\bibitem{classification}
J.~MACQUEEN, ``Some methods for classification and analysis of multivariate
  observations,'' \emph{Berkeley Symposium on Mathematical Statistics and
  Probability}, pp. 281--297, 1967.

\bibitem{Heuristic}
K.~W. Kintigh and A.~J. Ammerman, ``Heuristic approaches to spatial analysis in
  archaeology,'' \emph{American Antiquity}, vol.~47, no.~1, pp. 31--63, 1982.

\bibitem{kmeans}
T.~Kanungo, D.~M. Mount, N.~S. Netanyahu, C.~D. Piatko, R.~Silverman, and A.~Y.
  Wu, ``An efficient k-means clustering algorithm: Analysis and
  implementation,'' \emph{IEEE transactions on pattern analysis and machine
  intelligence}, vol.~24, no.~7, pp. 881--892, 2002.

\bibitem{mujoco}
E.~Todorov, T.~Erez, and Y.~Tassa, ``Mujoco: A physics engine for model-based
  control,'' \emph{2012 IEEE/RSJ International Conference on Intelligent Robots
  and Systems}, pp. 5026--5033, 2012.

\bibitem{benchmark_safe}
A.~Ray, J.~Achiam, and D.~Amodei, ``Benchmarking safe exploration in deep
  reinforcement learning,'' \emph{arXiv preprint arXiv:1910.01708}, 2019.

\bibitem{benchmark_gym}
G.~Brockman, V.~Cheung, L.~Pettersson, J.~Schneider, J.~Schulman, J.~Tang, and
  W.~Zaremba, ``Openai gym,'' \emph{arXiv preprint arXiv:1606.01540}, 2016.

\bibitem{comprehensive}
J.~Garcıa and F.~Fernández, ``A comprehensive survey on safe reinforcement
  learning,'' \emph{Machine Learning Research}, vol.~16, no.~1, pp. 1437--1480,
  2015.

\bibitem{cmdp}
E.~Altman, \emph{Constrained Markov decision processes}.\hskip 1em plus 0.5em
  minus 0.4em\relax CRC Press, 1999, vol. volume 7.

\bibitem{risk}
Y.~Chow, M.~Ghavamzadeh, L.~Janson, and M.~Pavone, ``Risk-constrained
  reinforcement learning with percentile risk criteria,'' \emph{Journal of
  Machine Learning Research}, vol.~18, pp. 6070--6120, 2017.

\bibitem{cmdpLag}
E.~Altman, ``Constrained markov decision processes with total cost criteria:
  Lagrangian approach and dual linear program,'' \emph{Mathematical methods of
  operations research}, vol.~48, pp. 387--417, 1998.

\bibitem{lyapunov}
Y.~Chow, O.~Nachum, A.~Faust, M.~Ghavamzadeh, and E.~Duenez-Guzman,
  ``Lyapunov-based safe policy optimization for continuous control,''
  \emph{arXiv preprint arXiv:1901.1003}, 2019.

\bibitem{cpo}
J.~Achiam, D.~Held, A.~Tamar, and P.~Abbeel, ``Constrained policy
  optimization,'' \emph{Proceedings of the International Conference on Machine
  Learning}, pp. 22--31, 2017.

\bibitem{trpo}
J.~Schulman, S.~Levine, P.~Abbeel, M.~Jordan, and P.~Moritz, ``Trust region
  policy optimization,'' \emph{Proceedings of the International Conference on
  Machine Learning}, pp. 1889--1897, 2015a.

\bibitem{barrier}
H.~Sikchi, W.~Zhou, and D.~Held, ``Lyapunov barrier policy optimization,''
  \emph{NeurIPS Deep Reinforcement Learning Workshop}, 2020.

\bibitem{projection}
T.-Y. Yang, J.~Rosca, K.~Narasimhan, and P.~J. Ramadge, ``Projection-based
  constrained policy optimization,'' \emph{International Conference on Learning
  Representations (ICLR)}, 2020.

\bibitem{accelerating}
------, ``Accelerating safe reinforcement learning with constraint-mismatched
  policies,'' \emph{International Conference on Machine Learning (ICML)}, 2021.

\bibitem{guarantee}
F.~Berkenkamp, M.~Turchetta, A.~Schoellig, and A.~Krause, ``Safe model-based
  reinforcement learning with stability guarantees,'' \emph{Advances in neural
  information processing systems}, pp. 908--918, 2017.

\bibitem{predictive}
T.~Koller, F.~Berkenkamp, M.~Turchetta, and A.~Krause, ``Learning-based model
  predictive control for safe exploration,'' \emph{2018 IEEE Conference on
  Decision and Control (CDC)}, pp. 6059--6066, 2018.

\bibitem{dalal2018}
G.~Dalal, K.~Dvijotham, M.~Vecerik, T.~Hester, C.~Paduraru, and Y.~Tassa,
  ``Safe exploration in continuous action spaces,'' \emph{arXiv preprint
  arXiv:1801.08757}, 2018.

\bibitem{dynamic}
L.~Zhu, Y.~Cui, and T.~Matsubara, ``Dynamic actor-advisor programming for
  scalable safe reinforcement learning,'' \emph{2020 IEEE International
  Conference on Robotics and Automation (ICRA)}, 2020.

\bibitem{curriculum}
M.~Turchetta, A.~Kolobov, S.~Shah, A.~Krause, and A.~Agarwal, ``Safe
  reinforcement learning via curriculum induction,'' \emph{arXiv preprint
  arXiv:2006.12136}, 2020.

\bibitem{autonomous}
C.~Ye, H.~Ma, X.~Zhang, K.~Zhang, and S.~You, ``Survival-oriented reinforcement
  learning model: An efficient and robust deep reinforcement learning algorithm
  for autonomous driving problem,'' \emph{International Conference on Image and
  Graphics}, pp. 417--429, 2017.

\bibitem{recovery}
B.~Thananjeyan, A.~Balakrishna, S.~Nair, M.~Luo, K.~Srinivasan, M.~Hwang, J.~E.
  Gonzalez, J.~Ibarz, C.~Finn, and K.~Goldberg, ``Recovery rl: Safe
  reinforcement learning with learned recovery zones,'' \emph{IEEE Robotics and
  Automation Letters}, vol.~6, no.~3, pp. 4915--4922, 2021.

\bibitem{trace}
B.~Eysenbach, S.~Gu, J.~Ibarz, and S.~Levine, ``Leave no trace: Learning to
  reset for safe and autonomous reinforcement learning,'' \emph{International
  Conference on Learning Representations (ICLR)}, 2018.

\bibitem{multi}
W.~Han, S.~Levine, and P.~Abbeel, ``Learning compound multi-step controllers
  under unknown dynamics,'' \emph{International Conference on Intelligent
  Robots and Systems (IROS)}, 2015.

\bibitem{PoseRBPF}
X.~Deng, A.~Mousavian, Y.~Xiang, F.~Xia, T.~Bretl, and D.~Fox, ``Poserbpf: A
  rao-blackwellized particle filter for 6d object pose tracking,''
  \emph{Robotics: Science and Systems(RSS)}, 2019.

\bibitem{ppo}
J.~Schulman, F.~Wolski, P.~Dhariwal, A.~Radford, and O.~Klimov, ``Proximal
  policy optimization algorithms,'' \emph{arXiv preprint arXiv:1707.06347},
  2017.

\end{thebibliography}
\bibliographystyle{IEEEtran}
\end{document}